\documentclass[10pt,twocolumn,letterpaper]{article}

\usepackage{iccv}
\usepackage{times}
\usepackage{epsfig}
\usepackage{graphicx}
\usepackage{amsmath}
\usepackage{amssymb}

\usepackage{multirow}
\usepackage{multicol}     
\usepackage[ruled]{algorithm2e}

\usepackage{algpseudocode}                                                                                                         

\usepackage[pagebackref=true,breaklinks=true,colorlinks,bookmarks=false]{hyperref}

\iccvfinalcopy %

\def\iccvPaperID{11073} %

\ificcvfinal\fi

\begin{document}

\def\ie{\emph{i.e.}\xspace}
\def\eg{\emph{e.g.}\xspace}
\def\wrt{\emph{w.r.t.}\xspace}

\newcommand{\Fig}[1]{Fig.~\ref{fig:#1}}
\newcommand{\Sec}[1]{Sec.~\ref{sec:#1}}
\newcommand{\Eq}[1]{Eq.~(\ref{eq:#1})}
\newcommand{\Tbl}[1]{Tab.~\ref{tab:#1}}
\newcommand{\Alg}[1]{Algorithm \ref{algo:#1}}
\newcommand{\ahyun}[1]{{\textcolor{red}{Ahyun: #1}}}
\newcommand{\woohyeon}[1]{{\textcolor{blue}{Woohyeon: #1}}}

\title{Learning to Discover Reflection Symmetry via Polar Matching Convolution}
\author{Ahyun Seo$^*$ \quad Woohyeon Shim$^*$ \quad Minsu Cho\vspace{0.15cm}\\
Pohang University of Science and Technology (POSTECH), South Korea\\
{\small \url{http://cvlab.postech.ac.kr/research/PMCNet}}
}

\maketitle
\def\thefootnote{*}\footnotetext{Equal contributions.} \def\thefootnote{\arabic{footnote}}
\ificcvfinal\thispagestyle{empty}\fi

\begin{abstract}
The task of reflection symmetry detection remains challenging due to significant variations and ambiguities of symmetry patterns in the wild.
Furthermore, since the local regions are required to match in reflection for detecting a symmetry pattern, it is hard for standard convolutional networks, which are not equivariant to rotation and reflection, to learn the task. 
To address the issue, we introduce a new convolutional technique, dubbed the polar matching convolution, which leverages a polar feature pooling, a self-similarity encoding, and a systematic kernel design for axes of different angles. 
The proposed high-dimensional kernel convolution network effectively learns to discover symmetry patterns from real-world images, overcoming the limitations of standard convolution. 
In addition, we present a new dataset and introduce a self-supervised learning strategy by augmenting the dataset with synthesizing images. 
Experiments demonstrate that our method outperforms state-of-the-art methods in terms of accuracy and robustness.
\end{abstract}

\section{Introduction}
\label{sec:introduction}

The world is built on symmetry. From the physical structures of nature, the biological patterns of life, to the artifacts of human, symmetries reveal themselves almost everywhere. Perception of such symmetries plays a crucial role at different levels of human vision~\cite{wagemans1995detection}; it provides humans with pre-attentive cues for early visual analysis and also acts as an integral part for 3D object perception under perspective distortion. 
Among common groups of symmetry, the most basic and popular form is reflection, mirror, or bilateral symmetry, which is the focus of this work. The task of reflection symmetry detection is to discover reflective patterns from images by detecting their axes of symmetry.
Despite the apparent simplicity of the mathematical concept~\cite{weyl1952symmetry} and the long history of research~\cite{liu2010computational}, the problem remains challenging due to significant variations and ambiguities of symmetry patterns in the wild. 

\begin{figure}[t]
    \centering
    \includegraphics[width=0.45\textwidth]{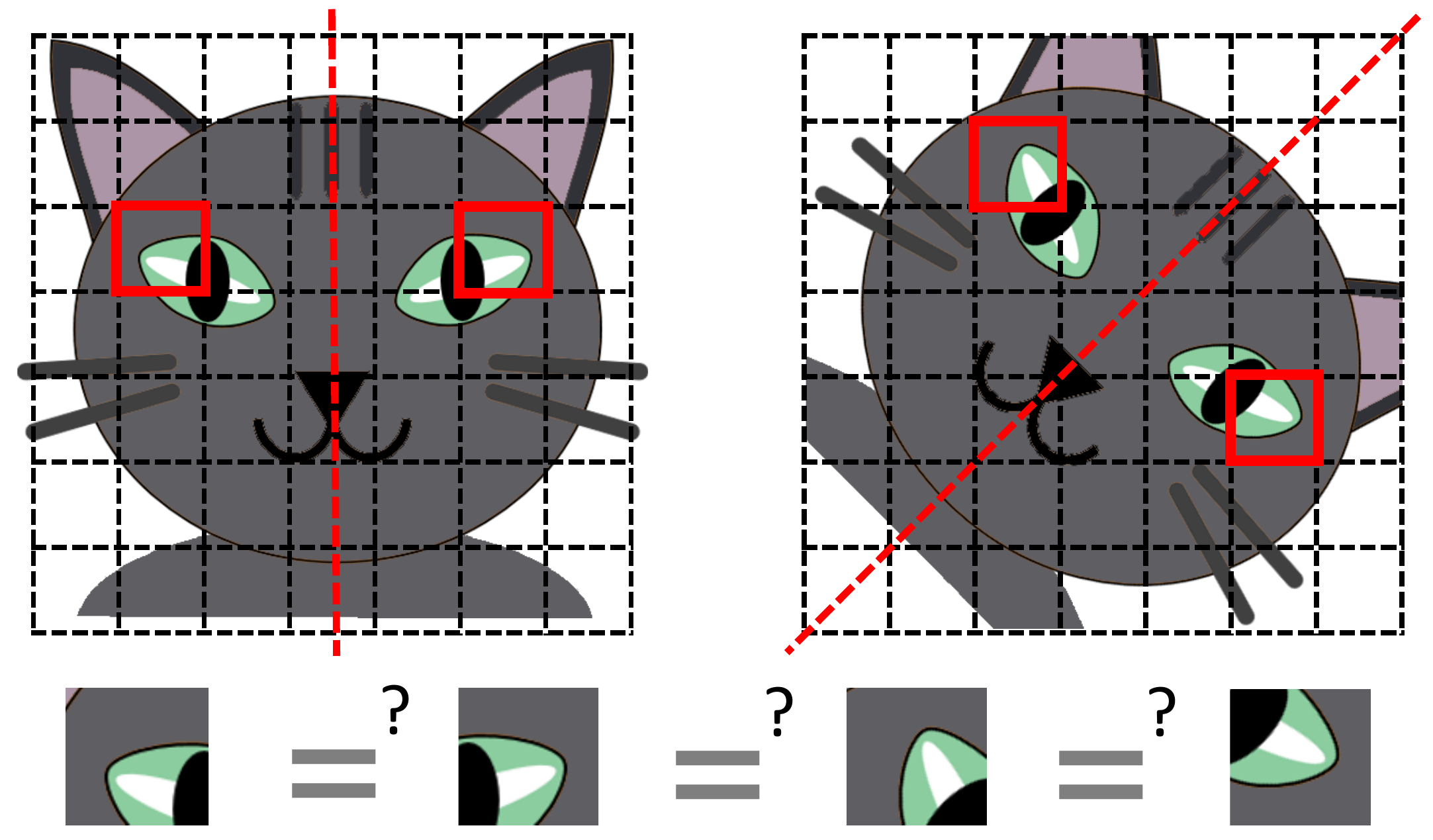}
    \caption{Feature matching in reflection symmetry. Perception of reflection symmetry requires matching features of corresponding regions in reflection with respect to its symmetry axis. Grids represent feature maps, and red squares denote corresponding regions.}\label{fig:teaser}
    \vspace{-2mm}
\end{figure}

One of the most promising directions for tackling the challenge would be to learn from data~\cite{halevy2009unreasonable}. 
While deep convolutional neural networks (CNNs) have made remarkable progress in a wide range of computer vision problems, there has been little work for learning symmetry detection on real-world images. 
Previous work~\cite{funk2017beyond} adapts a CNN~\cite{CP2016Deeplabv2} to regress a dense heatmap of symmetry axes for symmetry detection. 
While the results demonstrate the effectiveness of learning, the method does not consider the limitation of conventional CNNs in learning symmetry. 
As illustrated in Fig.~\ref{fig:teaser}, discovering symmetry requires recognizing matching pairs of local regions in reflection, and this may make it hard for CNNs, which are neither invariant nor equivariant to reflection~\cite{cohen2016group, kondor2018generalization}, to learn the task. 
Furthermore, the rotational freedom of the symmetry axis makes it even harder since conventional CNNs are not invariant to rotation either.
Standard convolution alone is not effective in learning the required properties solely from data.  

In this work, we introduce a new convolutional technique, dubbed the {\em polar matching convolution} (PMC), which leverages a polar feature pooling, a self-similarity encoding, and a systematic kernel design for axes of different angles.
On the basis of symmetrically matched feature pairs in a polar structure, the polar matching kernel computes the confidence of symmetry by exploiting both local similarities and geometric layouts of local patterns. The proposed convolution neural network learns to discover symmetry patterns with the high-dimensional kernel that effectively compares corresponding features in reflection with respect to the possible candidate axes of symmetry.
We also present a new symmetry detection dataset and introduce a self-supervised learning strategy using synthesized images. 
The experimental evaluation on the SDRW~\cite{liu2013symmetry} benchmark and our dataset shows that the proposed method, PMCNet, outperforms state-of-the-art methods in accuracy and robustness.

\section{Related Work}
\label{sec:relwork} 

Reflection symmetry algorithms can be broadly classified into two classes depending on whether the symmetry axes are detected using keypoint matching~\cite{cho2009bilateral, cicconet2017finding, loy2006detecting, nagar2017symmmap, wang2015reflection} or heatmap prediction~\cite{FUKUSHIMA.2004.08.001, FUKUSHIMA20061827, funk2017beyond, tsogkas2012learning}.

\paragraph{Keypoint matching.}
Early work on symmetry detection~\cite{cho2009bilateral,loy2006detecting} constructs the symmetry axes using the sparse set of symmetry matches obtained by matching the local features of the keypoints.
Keypoint matching is done by comparing the original features and their mirrored counterparts.
The descriptors need to be equivariant under reflection on images.
Loy and Eklundh~\cite{loy2006detecting} use the SIFT~\cite{lowe2004distinctive} descriptor so they can compute the reflected counterpart of the descriptor by reordering its elements.
For each pair, a candidate of a symmetry axis is the line perpendicularly passing through the mid-point of the symmetry pair.
Patraucean \etal~\cite{patraucean2013detection} develop a validation scheme using contrario theory on top of the work~\cite{loy2006detecting} to find the best mirror-symmetric image patches.
Cicconet \etal~\cite{cicconet2014mirror} propose a pairwise voting scheme based on the symmetry coefficient computed with the tangents. 
Elawady \etal~\cite{elawady2017multiple} transform the voting problem into the kernel density estimation and handle the displacement and orientation information with a linear-directional kernel-based voting representation.
Cho and Lee~\cite{cho2009bilateral} propose a symmetry-growing method to exploit further information than the local symmetry regions. The initial symmetry feature pairs are used as seeds to be merged and refined with geometric consistency and photometric similarity.
The methods~\cite{cicconet2017finding, nagar2017symmmap} group the pixel correspondences using a randomized algorithm and registration.
Cicconet \etal~\cite{cicconet2017finding} propose a Mirror Symmetry via Registration (MSR) framework to perform registration between the original and reflected patterns using normalized cross-correlation matches. They obtain the optimal symmetry plan with the reflection and registration mappings.
Cornelius \etal~\cite{cornelius2006detecting} consider detecting reflection symmetries on planar surfaces and locate them using a hough-voting scheme with feature quadruplets. They also improve this framework efficiently with the local affine frame \cite{cornelius2007efficient} that takes a single symmetry pair to hypothesize the symmetry axis.
Sinha \etal~\cite{sinha2012detecting} adopt a camera coordinate system to assume planar symmetry and use random sample consensus (RANSAC) algorithm to extract the camera parameters and the representative symmetry.
Our convolutional technique, PMC, is inspired by these keypoint matching schemes and designed to detect deformed real-world symmetries in the images.

\paragraph{Heatmap prediction.}

Other than comparing between features, recently proposed methods directly predict the symmetry for each pixel.
Tsogkas \etal~\cite{tsogkas2012learning} assign each pixel a bag of features at all orientations and scales to obtain the symmetry probability map. These symmetry structures are found with multiple instance learning.
Nagar \etal~\cite{nagar2017symmmap} leverage PatchMatch~\cite{barnes2009patchmatch} to produce dense SymmMap comprising the displacement field to the mirror reflection and the confidence score for each pixel.
SymmMap is first initialized with the sparse set of symmetry points and iteratively updated by searching neighbors which maximize the confidence scores.
Fukushima \etal~\cite{FUKUSHIMA.2004.08.001, FUKUSHIMA20061827} introduce a 4-layer neural network rather than using hand-crafted features. The network extracts and blurs the edges from the image to produce a dense heatmap of the symmetry axes.

Funk \etal~\cite{funk2017beyond} is the first to use deep CNN to regress the symmetry heatmap directly.
This work~\cite{funk2017beyond} employs a segmentation-based model~\cite{CP2016Deeplabv2} and trains with $\ell2$-loss over ground-truth heaptmaps.
While the simple CNN model outperforms the methods using the local feature descriptors, it often fails to discover real-world symmetries with deformed axes since conventional CNNs are neither invariant nor equivariant to reflection and rotation.
To mitigate these difficulties, we extend PMC to use the self-similarity descriptor for detecting symmetries.

\section{Proposed Approach}
\label{sec:method}
\begin{figure*}[t!]
    \centering
    \includegraphics[width=0.98\textwidth]{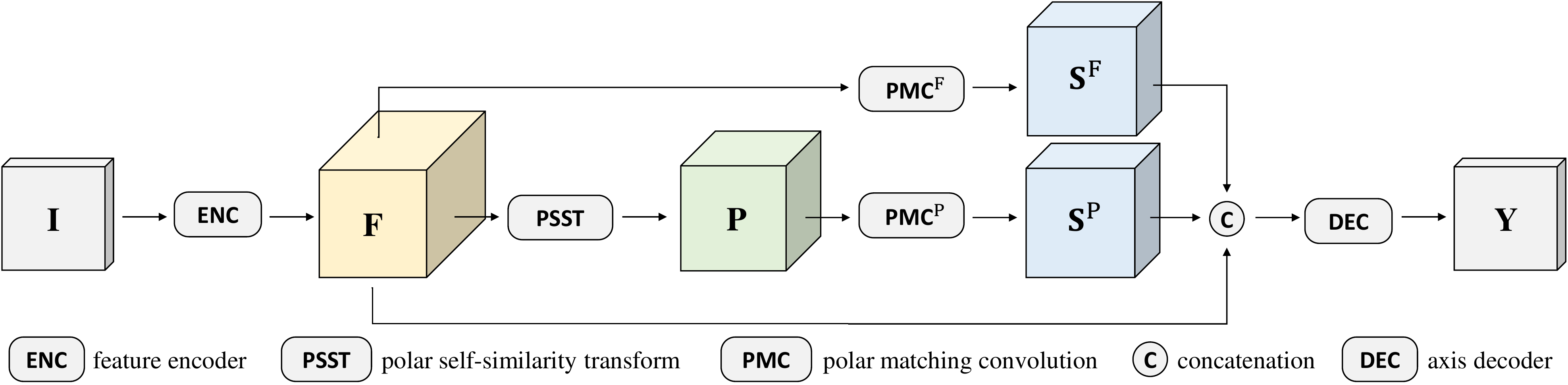}
    \caption{
   Overview of our proposed method. 
  For details, see \Sec{method}.
    }
    \label{fig:arch}
\end{figure*}

\begin{figure*}[t!]
    \centering
    \includegraphics[width=\textwidth]{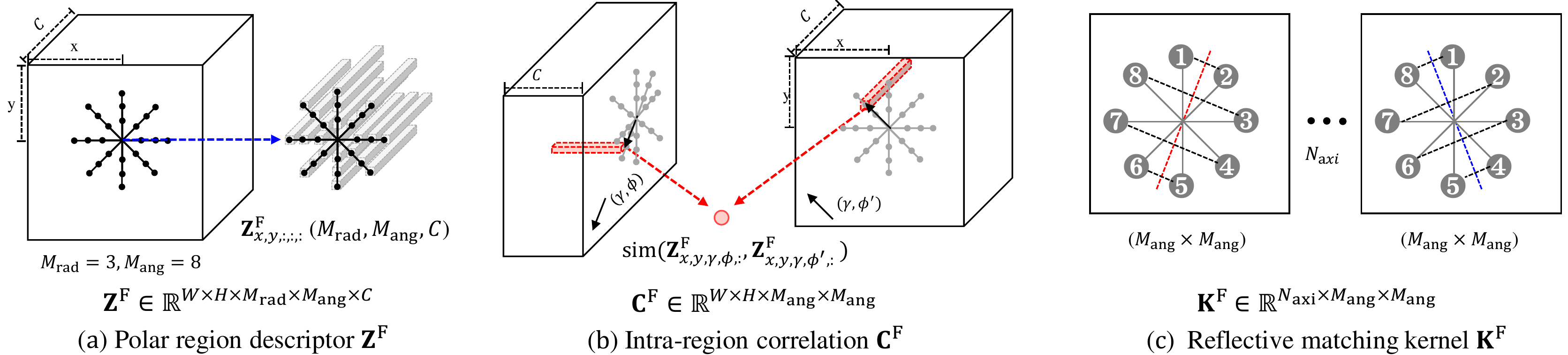}
    \caption{Illustration of the Polar Matching Convolution ($\text{PMC}^\mathrm{F}$).
    (a) The polar region descriptor $\mathbf{Z}^\mathrm{F}$ is sampled from the given base feature $\mathbf{F}$ with the number of sampling angles $M_\mathrm{ang}$ and radii $M_\mathrm{rad}$. 
    (b) The intra-region correlation $\mathbf{C}^\mathrm{F}$ are computed using the polar region descriptor $\mathbf{Z}^\mathrm{F}$. (c) The reflective matching kernel $\mathbf{K}^\mathrm{F}$ is applied to $\mathbf{C}^\mathrm{F}$ to compute the symmetry score tensor $\mathbf{S}^\mathrm{F}$. The matching feature pairs are indicated with black dotted lines.
    Note that $N_\mathrm{axi}$ is the number of the candidate axes.}
    \label{fig:prm}
\end{figure*}
The proposed method discovers reflection symmetry patterns by learning convolution with high-dimensional kernels for reflective feature matching.
Figure~\ref{fig:arch} briefly illustrates the overall architecture. 
Given an input image $\mathbf{I}$, a base feature $\mathbf{F}$ is computed by a feature encoder $\text{ENC}$. 
We transform the base feature $\mathbf{F}$ to a polar self-similarity descriptor $\mathbf{P}$.
Then, the polar matching convolutions, $\text{PMC}^\mathrm{F}$ and $\text{PMC}^\mathrm{P}$, compute the symmetry scores, $\mathbf{S}^\mathrm{F}$ and $\mathbf{S}^\mathrm{P}$, for the features, $\mathbf{F}$ and $\mathbf{P}$, respectively.
The final prediction is obtained by applying a convolutional decoder $\text{DEC}$ after combining the scores $\mathbf{S}^\mathrm{F}$ and $\mathbf{S}^\mathrm{P}$, and the base feature $\mathbf{F}$.
In the following, we elaborate the polar matching convolutions $\text{PMC}^\mathrm{F}$ and $\text{PMC}^\mathrm{P}$ and then describe the final output and the training objective for our model, the polar matching convolution network. 

\subsection{Polar Matching Convolution (PMC)}
\label{sec:polar-ref-match}
The convolution operation detects local patterns by multiplying shared kernels with features at different positions. %
The standard convolution in neural networks is trained to learn the kernels according to the target objective without specific constraints in convolution.  %
For effective symmetry detection, we propose the polar matching convolution~(PMC) that is designed to learn to extract symmetry patterns \wrt the axes of symmetry. The operations of PMC are illustrated in \Fig{prm} and detailed in the following. 

\paragraph{Polar region descriptor.}
As a basic unit for local matching, we use a polar region descriptor (\Fig{prm}a).
For each spatial position $(x, y)$ of feature tensor $\mathbf{F} \in \mathbb{R}^{W \times H \times C}$, we collect the features of the neighborhood points sampled with a polar grid centered at $(x, y)$.
We set the maximum radius as $M_\mathrm{len}$, and the number of sampling angles $M_\mathrm{ang}$ and radii $M_\mathrm{rad}$.  
A polar region descriptor $\mathbf{Z}^\mathrm{F} \in \mathbb{R}^{W \times H \times M_\mathrm{rad} \times M_\mathrm{ang} \times C}$ is designed to contain the local polar window for all spatial position $(x, y)$ sampled with bilinear interpolation from $\mathbf{F}$:  
\begin{align}
  \mathbf{Z}^\mathrm{F}_{x, y, \gamma, \phi, c}  = 
  \mathbf{F}_{x - \gamma \cos \phi, y + \gamma \sin \phi, c},
\end{align}
where $\phi$ indicates the offset angle and $\gamma$ offset radius from the center of each polar region.
The offsets $\phi$ and $\gamma$ are evenly distributed in $[0, \tfrac{M_\mathrm{len}}{M_\mathrm{rad}}]$ and $[0, \tfrac{2 \pi}{M_\mathrm{ang}}]$, respectively.

\begin{figure*}[t!]
    \centering
    \includegraphics[width=\textwidth]{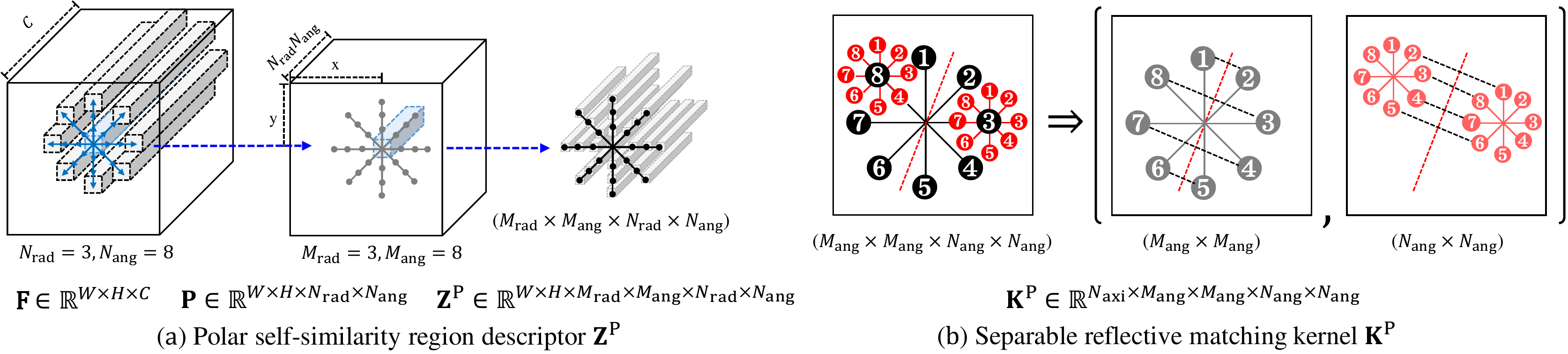}
    \caption{Illustration of PMC with self-similarity ($\text{PMC}^\mathrm{P}$).
    (a) The polar self-similarity $\mathbf{P}$ contains self-similarity values of the neighborhood pixels sampled with $N_\mathrm{rad}$ and $N_\mathrm{ang}$. The polar region descriptor $\mathbf{Z}^\mathrm{P}$ is then sampled from the polar self-similarity $\mathbf{P}$ with the sampling angle $M_\mathrm{ang}$ and radii $M_\mathrm{rad}$.
    (b) The reflective matching kernel $\mathbf{K}^\mathrm{P}$ extracts the relevant angle-wise relation pairs for $N_{\mathrm{axi}}$ candidate axes. 
    The kernel can be decomposed to the lower-dimensional kernels to tackle the relations within the polar region descriptor and the polar self-similarity descriptor.
    }
    \label{fig:conv}
\end{figure*}

\paragraph{Intra-region correlation.}
To capture the symmetry pattern within each polar region, we compute an intra-region correlation tensor $\mathbf{C}^\mathrm{F} \in \mathbb{R}^{W \times H \times M_\mathrm{ang} \times M_\mathrm{ang}}$ (\Fig{prm}b) as
\begin{align}
  \mathbf{C}^\mathrm{F}_{x, y, \phi, \phi'} = 
  \sum_{\gamma}
  \mathrm{sim}(\mathbf{Z}^\mathrm{F}_{x, y, \gamma, \phi, :}, 
  \mathbf{Z}^\mathrm{F}_{x, y, \gamma, \phi', :}),
\end{align}
to contain the similarities across different angles.
For the similarity function, the cosine-similarity is used. 
The colon (:) represents the whole elements in the specified axis.

\paragraph{Reflective matching kernel.}
From the intra-region correlation $\mathbf{C}^\mathrm{F}$, the symmetry axes are detected by the reflective matching kernel $\mathbf{K}^\mathrm{F} \in \mathbb{R}^{N_\mathrm{axi} \times M_\mathrm{ang} \times M_\mathrm{ang}}$ where $N_\mathrm{axi}$ is the number of the candidate axes.
The kernel associates with all the pairs in the polar region by construction.
As shown in \Fig{prm}c, if the region is symmetry with respect to the axis of the red-dotted line, the (unordered) feature pairs $\{(1, 2), (3, 8), (4, 7), (5, 6)\}$ should have high correlations.
On the other hand, for the axis of the blue-dotted line, the matching feature pairs should be $\{(1, 8), (2, 7), (3, 6), (4, 5)\}$.
Likewise, we establish a matching set for each candidate axis. 
In training, we learn the parameters of kernel $\mathbf{K}^\mathrm{F}$ involved in the matching sets only and discard the other entries.
This results in a kernel with $N_\mathrm{axi}M_\mathrm{ang}$ trainable parameters in total. The symmetry score $\mathbf{S}^\mathrm{F} \in \mathbb{R}^{W \times H \times N_\mathrm{axi}}$ is computed by convolution with the reflective matching kernel as
\begin{align}
  \mathbf{S}^\mathrm{F}_{x, y, k} = 
  \sum_{\phi, \phi'}
  \mathbf{K}^\mathrm{F}_{k, \phi, \phi'}
  \mathbf{C}^\mathrm{F}_{x, y, \phi, \phi'} .
\end{align}

\begin{figure}[t]
    \centering
    \includegraphics[width=0.43\textwidth]{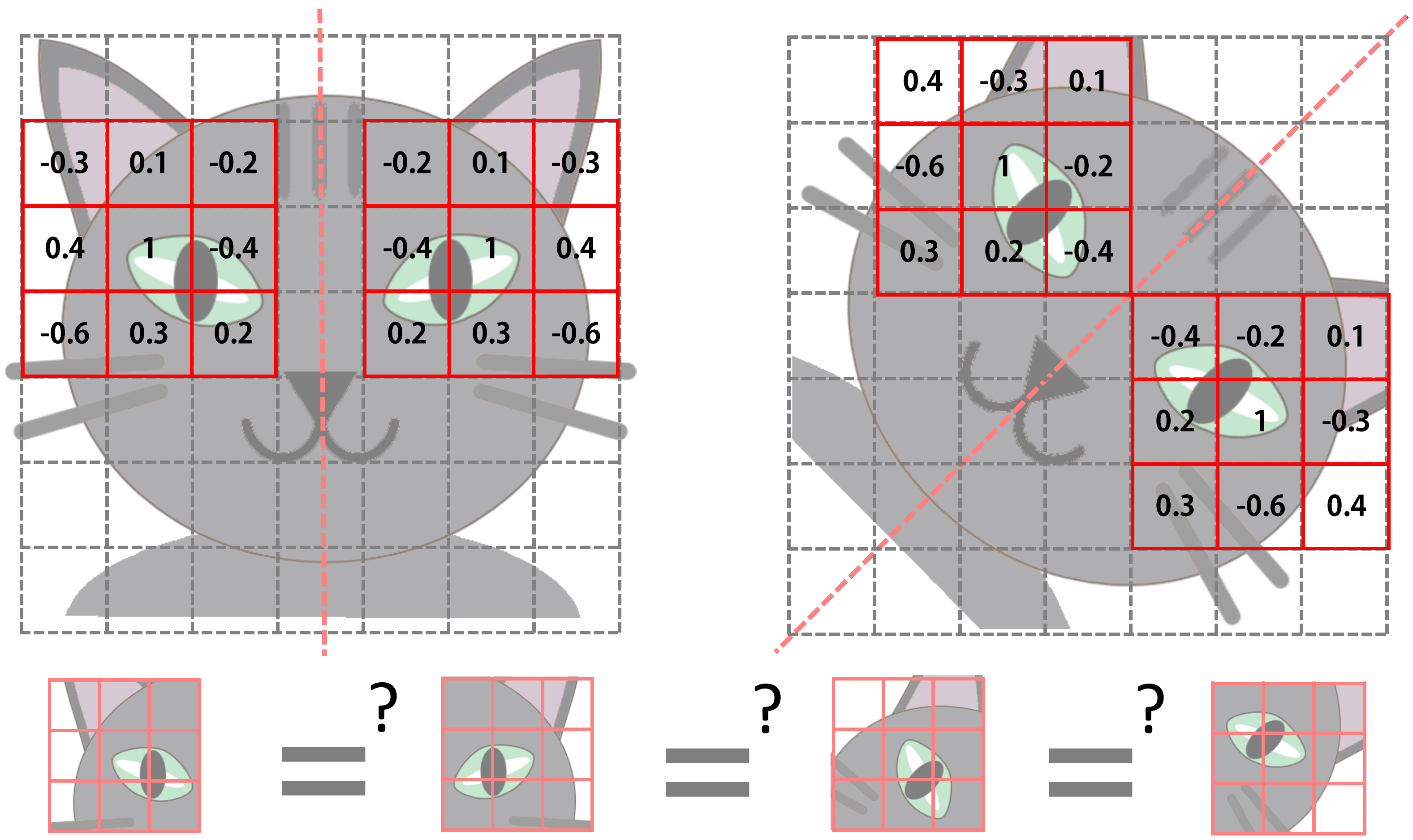}
    \caption{Matching with self-similarity in reflection symmetry. The red grid represents a self-similarity descriptor where each position encodes its similarity value to the center. Between the two corresponding regions in reflection, the self-similarity descriptor is preserved in reflection if each similarity value is invariant to reflection and rotation. See text for details. }
    \label{fig:pair_cat}
\end{figure}

\subsection{PMC with self-similarity}
\label{sec:pmc-selfsim}

Since CNNs are neither invariant nor equivariant to rotation and reflection, PMC may still have difficulty in learning to detect symmetry.   
For example, in the case of Fig.~\ref{fig:teaser}, PMC effectively detects reflection if the base feature is entirely invariant for both reflection and rotation. However, it may be an excessive requirement for the base feature extractor, \ie, lower convolutional layers, to achieve. 
To tackle this issue, we relax the requirement by extending PMC with the self-similarity descriptor~\cite{shechtman2007matching_selfsim}. 
Figure~\ref{fig:pair_cat} illustrates the basic idea. Let us represent each region as the descriptor of self-similarity values to the neighbor regions. The self-similarity descriptor is preserved in reflection if the pairwise similarity values are invariant to reflection and rotation. In other words, the region descriptor would be reflection-equivariant if similarity between two regions is invariant to reflection and rotation. Note that this invariance requirement on pairwise similarities is weaker than the original invariance requirement on individual features. 
To this end, we transform the base feature into self-similarity by using the neighborhood in a polar grid. 
The details of the extended $\text{PMC}^\mathrm{P}$ are illustrated in \Fig{conv} and described in the following.

\paragraph{Polar self-similarity descriptor.}
We adopt self-similarity with a polar-shaped local window to encode relational information with varying orientations.
Given an base feature $\mathbf{F} \in \mathbb{R}^{W \times H \times C}$, polar self-similarity $\mathbf{P} \in \mathbb{R}^{W \times H \times N_\mathrm{rad} \times N_\mathrm{ang}}$ is computed as 
\begin{align}
  \mathbf{P}_{x, y, r, \theta} &=  \mathrm{sim} (\mathbf{F}_{x,y,:}, \mathbf{F}_{x - r \cos \theta, y+r \sin \theta, :}),
\end{align}
where $x$ and $y$ indicate the spatial position, and $r$ and $\theta$ denote the offsets for a spatial position in polar coordinate. $r$ and $\theta$ are evenly distributed between $[0, \tfrac{N_\mathrm{len}}{N_\mathrm{rad}}]$ and $[0, \tfrac{2 \pi}{N_\mathrm{ang}}]$, respectively, where $N_\mathrm{rad}$ is the number of sampled radii, $N_\mathrm{ang}$ the number of angles, and $N_\mathrm{len}$ the maximum radius. 
For the similarity function, the rectified cosine-similarity, which is the cosine similarity followed by ReLU. 
The polar region descriptor $\mathbf{Z}^\mathrm{P} \in \mathbb{R}^{W \times H \times M_\mathrm{rad} \times M_\mathrm{ang} \times N_\mathrm{rad} \times N_\mathrm{ang}}$ is constructed by sampling the polar self-similarity descriptor $\mathbf{P}$ with bilinear interpolation:
\begin{align}
  \mathbf{Z}^\mathrm{P}_{x, y, \gamma, \phi, r, \theta}  = 
  \mathbf{P}_{x - \gamma \cos \phi, y + \gamma \sin \phi, r, \theta}.
\end{align}
Note that we preserve the structure of the polar-shaped local windows of $\mathbf{P}$ so that we construct the match \wrt the polar self-similarity descriptor $\mathbf{P}$. 

\paragraph{Intra-region correlation.}
From the polar region descriptor $\mathbf{Z}^\mathrm{P}$, the intra-region correlation tensor 
$\mathbf{C}^\mathrm{P} \in \mathbb{R}^{W \times H \times M_\mathrm{ang} \times M_\mathrm{ang} \times N_\mathrm{ang} \times N_\mathrm{ang}}$ is computed as
\begin{align}
  \mathbf{C}^\mathrm{P}_{x, y, \phi, \phi', \theta, \theta'} = 
  \sum_{\gamma, r}
  \mathbf{Z}^\mathrm{P}_{x, y, \gamma, \phi, r, \theta}
  \mathbf{Z}^\mathrm{P}_{x, y, \gamma, \phi', r, \theta'}.
\end{align}
The operation above is equivalent to the outer product of $N_\mathrm{ang}$-dimensional polar self-similarity vectors in polar region, results in $(N_\mathrm{ang}\times N_\mathrm{ang})$ element-wise relations of the polar self-similarity descriptor $\mathbf{P}$. 

\paragraph{Separable reflective matching kernel.}
To detect symmetry axes from the intra-region correlation $\mathbf{C}^\mathrm{P}$, the reflective matching kernel $\mathbf{K}^\mathrm{P} \in \mathbb{R}^{N_\mathrm{axi} \times M_\mathrm{ang} \times M_\mathrm{ang} \times N_\mathrm{ang} \times N_\mathrm{ang}}$ is required. 
The symmetry score $\mathbf{S}^\mathrm{P} \in \mathbb{R}^{W \times H \times N_\mathrm{axi}}$ is computed by convolution with the reflective matching kernel as
\begin{align}
  \mathbf{S}^\mathrm{P}_{x, y, k} &= 
  \sum_{\phi, \phi', \theta, \theta'} 
  \mathbf{K}^\mathrm{P}_{k, \phi, \phi', \theta, \theta'}
  \mathbf{C}^\mathrm{P}_{x, y, \phi, \phi', \theta, \theta'}. 
  \label{eq:score_p}
\end{align}
Alternatively, the kernel $\mathbf{K}^\mathrm{P}$ can be computed by sequentially applying the kernels 
$\mathbf{K}^{\mathrm{P}_\mathrm{M}} \in \mathbb{R}^{N_\mathrm{axi} \times M_\mathrm{ang} \times M_\mathrm{ang}}$ and $\mathbf{K}^{\mathrm{P}_\mathrm{N}} \in \mathbb{R}^{N_\mathrm{axi} \times N_\mathrm{ang} \times N_\mathrm{ang}}$
as illustrated in \Fig{conv}b.
If the image is symmetry \wrt the axis of the red-dotted line, the (unordered) feature pairs $\{(1, 2), (3, 8), (4, 7), (5, 6)\}$ and the element pairs $\{(1, 2), (3, 8), (4, 7), (5, 6)\}$ should be matched. 
Different from \Sec{polar-ref-match}, the matching set of $\text{PMC}^\mathrm{P}$ emcompasses two levels of matching pairs.
When $M_\mathrm{ang}=N_\mathrm{ang}$, the self-similar pattern enables the kernels $\mathbf{K}^{\mathrm{P}_\mathrm{M}}$ and $\mathbf{K}^{\mathrm{P}_\mathrm{N}}$ to share the kernel parameters.
We denote the shared kernel by $\mathbf{K}^{\mathrm{P}_\mathrm{S}} \in \mathbb{R}^{N_\mathrm{axi} \times N_\mathrm{ang} \times N_\mathrm{ang}}$.
As in the reflective matching kernel of \Sec{polar-ref-match}, we only use and train $N_\mathrm{axi}N_\mathrm{ang}$ elements among $N_\mathrm{axi}N_\mathrm{ang}N_\mathrm{ang}$ parameters of the kernel.
We can thus rewrite \Eq{score_p} as
\begin{align}
  \Tilde{\mathbf{C}}^\mathrm{P}_{x, y, \theta, \theta'} &= \sum_{\phi, \phi'}
  \mathbf{K}^{\mathrm{P}_\mathrm{S}}_{k, \phi, \phi'}
  \mathbf{C}^\mathrm{P}_{x, y, \phi, \phi', \theta, \theta'} , \\
 \mathbf{S}^\mathrm{P}_{x, y, k} &= \sum_{\theta, \theta'} 
  \mathbf{K}^{\mathrm{P}_\mathrm{S}}_{k, \theta, \theta'}
  \Tilde{\mathbf{C}}^\mathrm{P}_{x, y, \theta, \theta'}.
\end{align}
\subsection{Reflection symmetry detection}

\label{sec:symdet}
We combine $\mathbf{S}^\mathrm{P}$, $\mathbf{S}^\mathrm{F}$, and $\mathbf{F}$ as the input to the decoder ($\text{DEC}$).
To feed the positional confidence to the decoder as an additional input, we construct the maximum score map  $\Tilde{\mathbf{S}} \in \mathbb{R}^{W \times H \times 1}$ by pooling the maximum score for each spatial position and  
then replace $\mathbf{S}$ with $[\Tilde{\mathbf{S}} || \mathbf{S}]$. 
The final prediction $\mathbf{Y}$ is the output of the decoder $g(\cdot)$:  
\begin{align}
  \mathbf{Y} &= g([
  \mathbf{S}^\mathrm{P} || \mathbf{S}^\mathrm{F} || \mathbf{F}]) ,
\end{align}
where $||$ denotes the concatenation operation along the channel. 
The decoder $g(\cdot)$ consists of two blocks of
$(3 \times 3 \ \mathrm{conv} \text{-} \mathrm{BN} \text{-} \mathrm{DO} \text{-} \mathrm{ReLU})$ and a $1 \times 1$ $\mathrm{conv}$ layer with sigmoid, where BN and DO denote batch normalization and dropout, respectively.

\subsection{Training objective}
Our network basically performs pixel-wise binary classification for detection of symmetry axes.
Since the symmetry axes occupy only a tiny part of the image, we adopt $\alpha$-variant of the focal loss \cite{lin2017focal} to down-weight the non-axes region in training:
\begin{align}
    \mathcal{L} &= \sum_{x,y} -\alpha_t (x, y) (1 - p_t (x, y))^{\beta} \log(p_t (x, y)),
\end{align}
where $\alpha_t (x,y)$ balances the importance of axes/non-axes region and $p_t (x,y)$ is a confidence of symmetry, which adjust the rate between the easy and the hard samples with the focusing parameter $\beta$~\cite{lin2017focal}. Moreover, we soft-weight $p_t (x, y)$ and $\alpha_t (x, y)$ after smoothing the corresponding label with gaussian blur with radius 5 following \cite{funk2017beyond},
\begin{align}
    p_t (x, y) &= \mathbf{Y}_{x,y} \mathbf{M}_{x,y} + (1 - \mathbf{Y}_{x,y})(1 - \mathbf{M}_{x,y}), \\
    \alpha_t (x, y) &= \alpha \mathbf{M}_{x,y} + (1 - \alpha)(1-\mathbf{M}_{x,y}),
\end{align}
where $\mathbf{Y}$ and $\mathbf{M}$ are the predicted heatmap and the ground-truth (GT) label heatmap and $\alpha$ is a pre-specified scalar that typically set by inverse class frequency.

\section{Experiments}

\begin{figure}[t]
    \centering
    \includegraphics[width=0.44\textwidth]{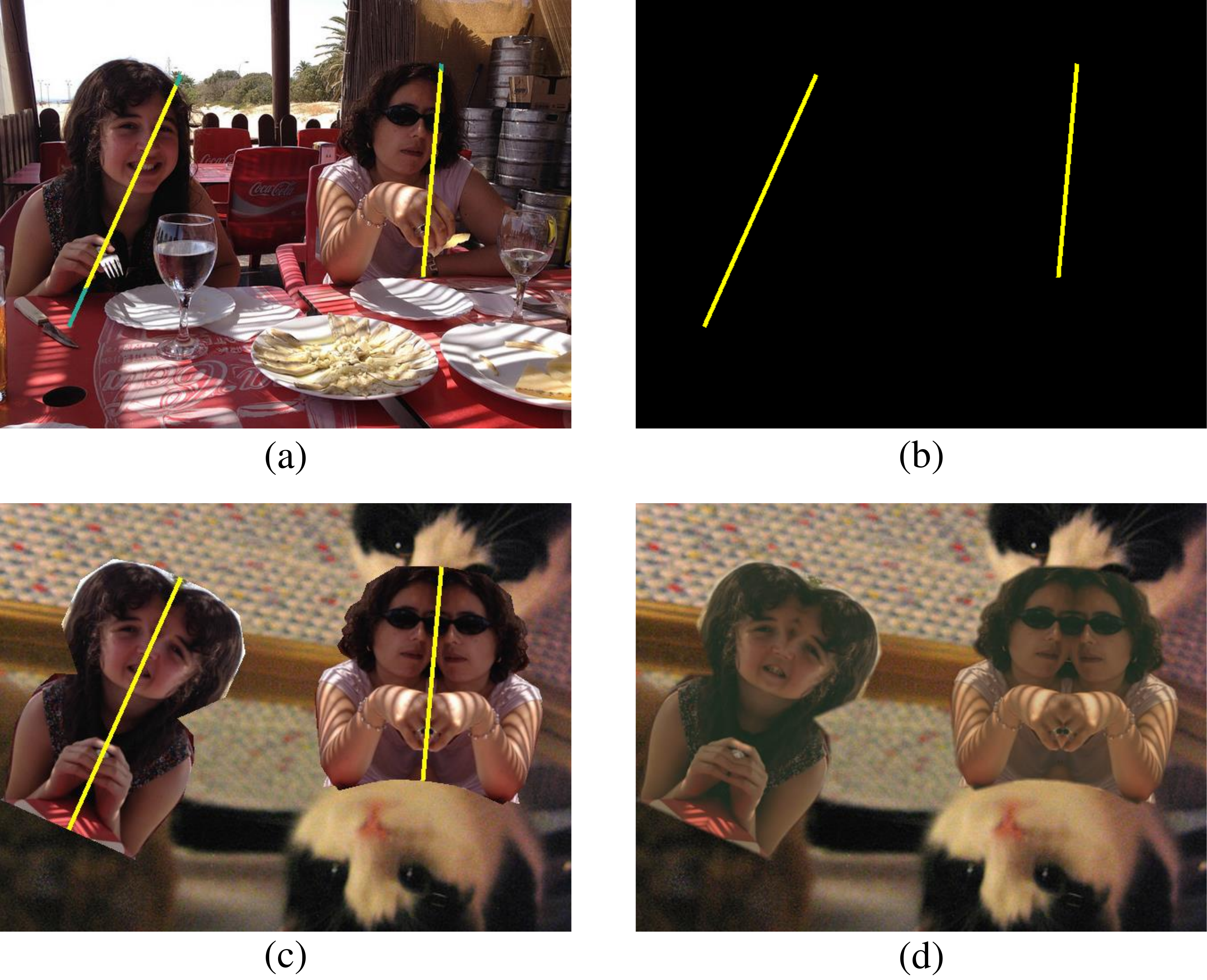}
    \caption{An example of a synthesized image. (a) A foreground image with randomly sampled axes. (b) Ground-truth symmetry axes. (c) The result of simple paste of the mirror-flipped foreground instances to the background image. (d) A final training sample with a blending procedure.}
    \label{fig:synth_ex}
\end{figure}

\subsection{Datasets}
\paragraph{Existing datasets.}
SDRW~\cite{liu2013symmetry} is used in \textit{Symmetry Detection from Real World Images Competition '13} which contains 51/70 images for training and testing purposes.
NYU~\cite{ConvSymm2016} consists of 176 single-symmetry and 63 multiple-symmetry images.
DSW~\cite{funk20172017} contains 100 images for each single and multiple symmetries and is used in the challenge \textit{Detecting Symmetry in the Wild '17}.
Sym-COCO~\cite{funk20172017} is the dataset built on images selected from COCO~\cite{lin2014microsoft} containing 250 training and 240 test images for reflection symmetry. 
BPS~\cite{funk2017beyond} is a dataset with 1,202 images collected from COCO to train the deep neural networks. The test set of Sym-COCO and BPS are the same.
Unfortunately, both DSW and BPS datasets are not available at this point. 
Therefore, we use the two available datasets, SDRW and NYU, 
and construct a new dataset, LDRS, and its synthetic augmentation, LDRS-\textit{synth}, as follows.

\paragraph{Our dataset (LDRS).}
We construct the LDRS dataset, which is named after the acronym of our work, following the dataset collection protocol of BPS~\cite{funk2017beyond} in terms of the scale and the characteristics of the images.
First, we filter the images from COCO~\cite{lin2014microsoft} dataset by excluding the instances with the area less than $\frac{hw}{16}$ given image height $h$ and width $w$ and discarding the images with less than three instances.
Among them, we manually select 1,500 candidate images with the valid reflection symmetries for the train/val split. 
For the test split, we selected 240 images from the 250 training images of Sym-COCO~\cite{funk20172017} as they were part of the test split of the COCO dataset.
We mainly follow the annotation guidelines of the previous datasets~\cite{ConvSymm2016, liu2013symmetry, funk2017beyond} except we do not consider the 3D prior-based or semantic symmetries.
Four human annotators labeled the symmetry axes as line segments using the annotation tool labelMe~\cite{labelme2016}.
The train/val/test split contains 1,110/127/240 images, respectively, and all of the images contain at least one ground-truth symmetry axis.

\paragraph{Synthetic augmentation.}
To overcome the limited real-world training data, we adopt a self-supervised learning strategy of using synthesized images from the existing dataset.
We use COCO~\cite{lin2014microsoft} images and annotations utilized for instance segmentation.
The images are generated online during training, and we denote this set of images as LDRS-\textit{synth}.
We pre-select the valid foreground images (6,526) while using the whole training split (83,000) for the background.
An example is shown in~\Fig{synth_ex}.
The synthesis process takes a pair of images, one for the foreground and the other for the background.
Given a foreground image and its mask, we select the top-5 instances by area.
For each instance, we randomly select two angles $\psi_1$ and $\psi_2$ to rotate the masks.
We assign vertical axes for each instance and mirror one of the two parts of the target instance.
Then, we rotate the image reverse so that each axis results in the angle of $-\psi_1$ and $-\psi_2$.
Among ten candidate instances, the combinations with the largest sum of the length of the axes are selected for foreground instances.
The instances are blurred with a radius of 5 and matched with the statistics of the background image after normalization for blending.
The detailed algorithm is stated in the supplementary material.

\begin{figure}[t]
    \centering
    \includegraphics[width=0.45\textwidth]{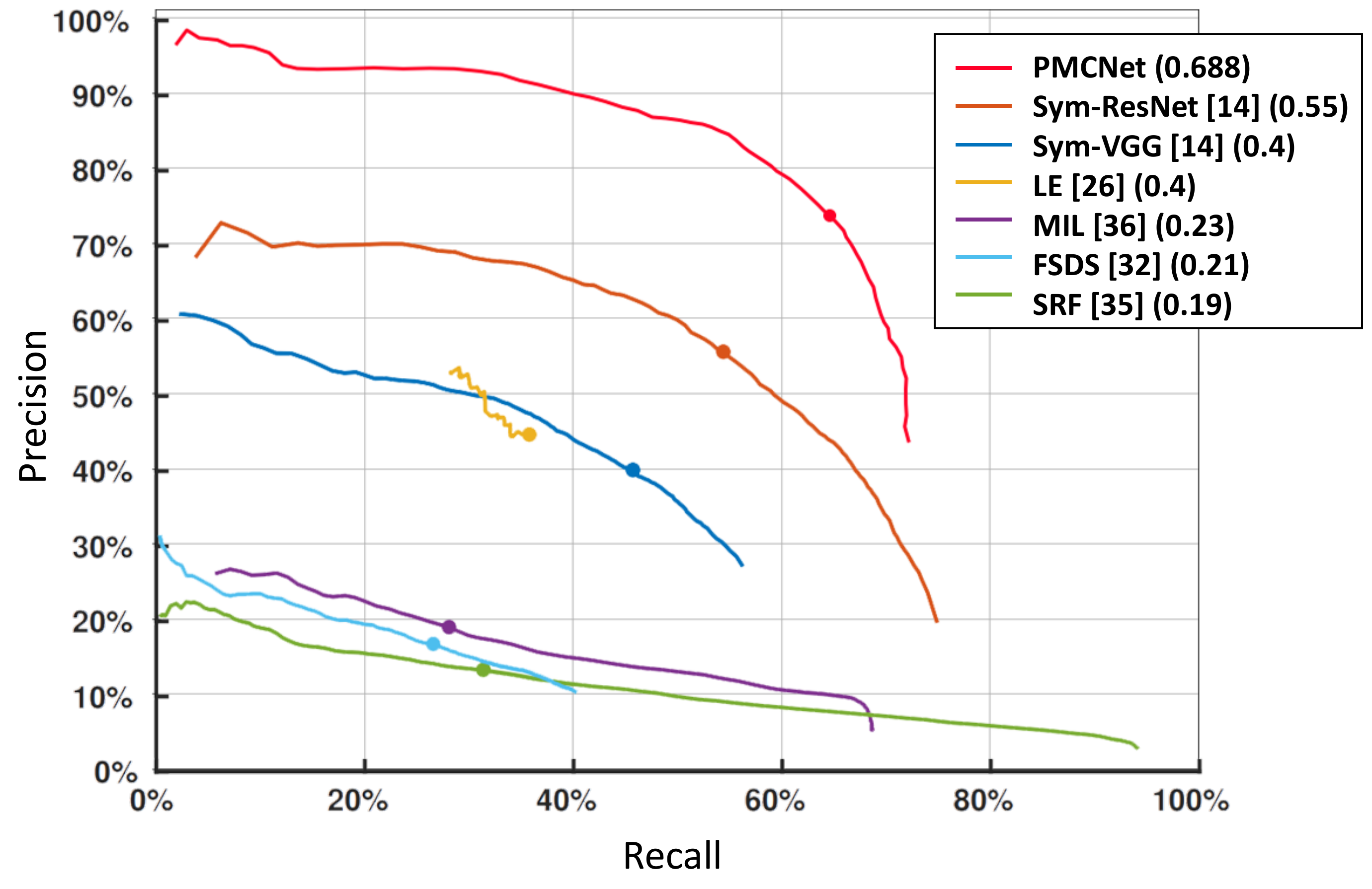}
    \caption{Precision-Recall curve on the SDRW~\cite{liu2013symmetry} symmetry detection dataset. The (recall, precision) point of the maximum F1-score is indicated with dots.}
    \label{fig:pr_curve}
\end{figure}

\subsection{Experimental setup}

\paragraph{Training and testing data.}
For training, we use the training splits of SDRW and LDRS along with the whole sets of NYU and LDRS-\textit{synth}.
For evaluation, we use the test splits of SDRW and LDRS.

\paragraph{Evaluation metric.}
Since the ground-truth axes are pixel-width, we follow the common procedure for evaluating the boundaries by applying the morphological thinning~\cite{elawady2017multiple, martin2004learning} to the output heatmap and performing the pixel matching algorithm between the output and the ground-truth boundaries. 
We set the maximum distance as 0.01 of the image diagonal for the pixel matching process.
We use 100 evenly distributed thresholds to compute the F1 score, the harmonic mean of precision and recall, following~\cite{elawady2017multiple}.
We report the maximum F1 score (\%) for the test split of both SDRW and our LDRS datasets for ablation studies.

\begin{table*}
\centering
\setlength\tabcolsep{4pt}
\begin{minipage}{0.36\textwidth}
\centering
\resizebox{0.94\textwidth}{!}{%
\begin{tabular}{|cc|cc|} \hline
\multicolumn{2}{|c|}{descriptor} & \multicolumn{2}{c|}{test dataset}  \\ \hline
polar region & polar self-similarity & SDRW & LDRS \\ \hline \hline
\checkmark  &           & 65.5 & 35.3 \\
            & \checkmark & 66.2 & 33.7 \\
\hline
\checkmark  & \checkmark & \textbf{68.3} & \textbf{35.9} \\
\hline
\end{tabular} 
}
\vspace{2mm}
 \caption{Ablation on descriptors.} 
 \label{tab:ablation-desc} 
\end{minipage}
\hfill
\begin{minipage}{0.27\textwidth}
\centering
\resizebox{0.94\textwidth}{!}{%
\begin{tabular}{|cc|cc|} \hline
\multicolumn{2}{|c|}{kernel} & \multicolumn{2}{c|}{test dataset}  \\ \hline
learnable & axis-aware &  SDRW & LDRS \\ \hline \hline
            & \checkmark  & 65.4 & 34.6 \\ 
\checkmark  &  & 67.5 & 35.3 \\ 
\hline
\checkmark  & \checkmark & \textbf{68.8} & \textbf{37.3} \\ 
\hline
\end{tabular} 
}
 \vspace{2mm}
 \caption{Ablation on kernels.} 
 \label{tab:ablation-kernel} 
\end{minipage}
\hfill
\begin{minipage}{0.34\textwidth}
\centering
\resizebox{0.94\textwidth}{!}{%
\begin{tabular}{|cc|cc|} \hline
\multicolumn{2}{|c|}{decoder input} & \multicolumn{2}{c|}{test dataset}  \\ \hline
 score ($\mathbf{S}^\mathrm{P}$, $\mathbf{S}^\mathrm{F}$) & base feature ($\mathbf{F}$) & SDRW & LDRS \\ \hline \hline
\checkmark &  & 62.7 & 32.5 \\
             &  \checkmark  & 66.8 & 36.2 \\
 \hline
 \checkmark  & \checkmark & \textbf{68.8} & \textbf{37.3} \\ \hline
\end{tabular} 
}
 \vspace{2mm}
 \caption{Ablation on decoder inputs.} 
 \label{tab:ablation-pred} 
\end{minipage}
\end{table*}

\begin{table*}
\centering
\setlength\tabcolsep{4pt}
\begin{minipage}{0.34\textwidth}
\centering
\resizebox{0.9\textwidth}{!}{%
\begin{tabular}{|cc|cc|} \hline
 \multicolumn{2}{|c|}{training strategy} & \multicolumn{2}{c|}{test dataset}  \\ \hline
 pre-training & fine-tuning & SDRW & LDRS \\ \hline \hline
 & \checkmark & 47.8 & 25.5 \\
\checkmark &  & 67.2 & 32.7 \\ \hline
\checkmark & \checkmark & \textbf{68.8} & \textbf{37.3} \\ \hline
\end{tabular} 
}
\vspace{2mm}
\caption{Ablation on training strategies.}
 \vspace{-2mm}
\label{tab:ablation-train}
\end{minipage}
\begin{minipage}{0.45\textwidth}
\centering
\resizebox{0.9\textwidth}{!}{%
\begin{tabular}{|cc|cc|} \hline
 \multicolumn{2}{|c|}{train dataset} & \multicolumn{2}{c|}{test dataset}  \\  \hline
 LDRS-\textit{synth} & SDRW + LDRS + NYU 
 & SDRW & LDRS  \\ \hline \hline
 \checkmark &  & 37.6 & 14.2 \\ 
  & \checkmark & 61.6 & 34.8 \\
\hline
\checkmark & \checkmark & \textbf{68.8} & \textbf{37.3} \\ \hline
\end{tabular} 
}
\vspace{2mm}
\caption{Ablation on training datasets.}
 \vspace{-2mm}
\label{tab:ablation-data}
\end{minipage}
\end{table*}

\paragraph{Implementation details.}
We employ ResNet-101~\cite{He_2016_CVPR} pre-trained on ImageNet~\cite{imagenet_cvpr09} as the backbone for the base feature $\mathbf{F}$. 
To aggregate multi-resolution features,
we apply the Atrous Spatial Pyramid Pooling module (ASPP)~\cite{DBLP:journals/corr/ChenPSA17}. %
The dropout~\cite{JMLR:v15:srivastava14a} rates of two blocks in the decoder $g(\cdot)$ are 0.5 and 0.1, respectively.
The hyperparameters for the polar descriptors are set as $M_\mathrm{rad} \text{=} 5$, $M_\mathrm{ang} \text{=} 8$, $N_\mathrm{rad} \text{=} 5$, $N_\mathrm{ang} \text{=} 8$, $M_\mathrm{len} \text{=} 8$, $N_\mathrm{len} \text{=} 4$, and $N_\mathrm{axi} \text{=} 8$. 
When $N_\mathrm{axi} \text{=} 8$, the 4 axes lie on the polar descriptor, and the other 4 axes lie in between the polar descriptor.
The focal loss is applied with the weighting factor $\alpha \text{=} 0.95$ and the focusing parameter $\beta \text{=} 2$. 
The model is trained end-to-end for 100 epochs with an initial learning rate $0.001$ using Adam optimizer~\cite{Adamsolver} while the learning rate is decayed by the factor of 0.1 at 50$^{\mathrm{th}}$ and 75$^{\mathrm{th}}$ epochs.
The training batch is randomly composed of 16 images from the different datasets.
The images are augmented with random rotation and color jittering.
During training, we evaluate our model every 5 epochs in our validation set with 10 thresholds.
The input images are resized to (417, 417) for both training and testing, while we resize the output back to its original size for evaluation.
We use PyTorch~\cite{NEURIPS2019_9015} framework to implement our model.

\subsection{Ablation study}
\paragraph{Different types of descriptors.}

We introduce two types of polar descriptors in \Sec{pmc-selfsim}: the \textit{polar region} descriptor and the \textit{polar self-similarity} descriptor.
While the $\text{PMC}^\mathrm{F}$ extracts the \textit{polar region} descriptor from the base feature, the $\text{PMC}^\mathrm{P}$ first computes the \textit{polar self-similarity} descriptor using the base feature, then constructs the \textit{polar region} descriptor.
To investigate the impact of the proposed polar descriptors, we conduct experiments using decoder input of $(\mathbf{S}||\mathbf{F})$, which contains a single score computed by the corresponding $\text{PMC}$. 
Without the \textit{polar region} descriptor, the region is a single pixel. 
As shown in \Tbl{ablation-desc}, using both polar descriptors are effective.

\paragraph{Reflective matching kernels.}
We show the effectiveness of the kernel design in PMC in \Tbl{ablation-kernel}. 
If the kernel is not \textit{axis-aware}, we use all parameters consisting of the kernels. 
If the kernel is not \textit{learnable}, we fix the values and do not update during training.
The \textit{learnable} and \textit{axis-aware} kernel results in the best performance.
Interestingly, the \textit{axis-aware} kernel increases the max F1 score while using only $\tfrac{1}{N_\mathrm{axi}}$ of the total learnable kernel parameters.

\paragraph{Decoder inputs.}
We compare the different inputs to the decoder of our model in \Tbl{ablation-pred}. 
The \textit{score} denotes the combination of $\mathbf{S}^\mathrm{P}$ and $\mathbf{S}^\mathrm{F}$ and the \textit{base feature} indicates $\mathbf{F}$.
Without the \textit{base feature}, the model scores lower than the final model because of the lack of semantic information from the CNN features.
The model with \textit{base feature} becomes a simple segmentation network similar to \cite{funk2017beyond}.
Since the \textit{base feature} does not contain the axis-aware pixel matching information, using \textit{score} improves the max F1 score. 

\paragraph{Training strategies.}
The backbone network of our model is initialized with the ImageNet pre-trained weights and fine-tuned during training. 
~\Tbl{ablation-train} shows comparisons to results (1) w/o pre-training on the ImageNet and (2) w/o fine-tuning the feature extractor.
The performance drops significantly without fine-tuning, but not in the case without pre-training.
While adapting the whole network to polar matching kernels is crucial, the pre-trained weights still act as a good initial prior for computing pixel-level similarities.

\paragraph{Training datasets.}
We investigate the effectiveness of the self-supervised learning strategy of using synthetically augmented images using the different combinations of the train datasets in ~\Tbl{ablation-data}. 
For the real-image datasets, we combine SDRW-\textit{train}~\cite{liu2013symmetry}, NYU~\cite{ConvSymm2016}, and LDRS-\textit{train} each consisting of 51, 239 and 1,110 images, respectively.
For the augmented images, we use LDRS-\textit{synth}, which is generated using COCO images. 
Training only with LDRS-\textit{synth} suffers from poor generalization ability, but it is helpful when used together with the real-image datasets as when the amount of the real images is far less than that of the augmented images.

\begin{figure*}[t!]
    \centering
    \includegraphics[width=0.96\textwidth]{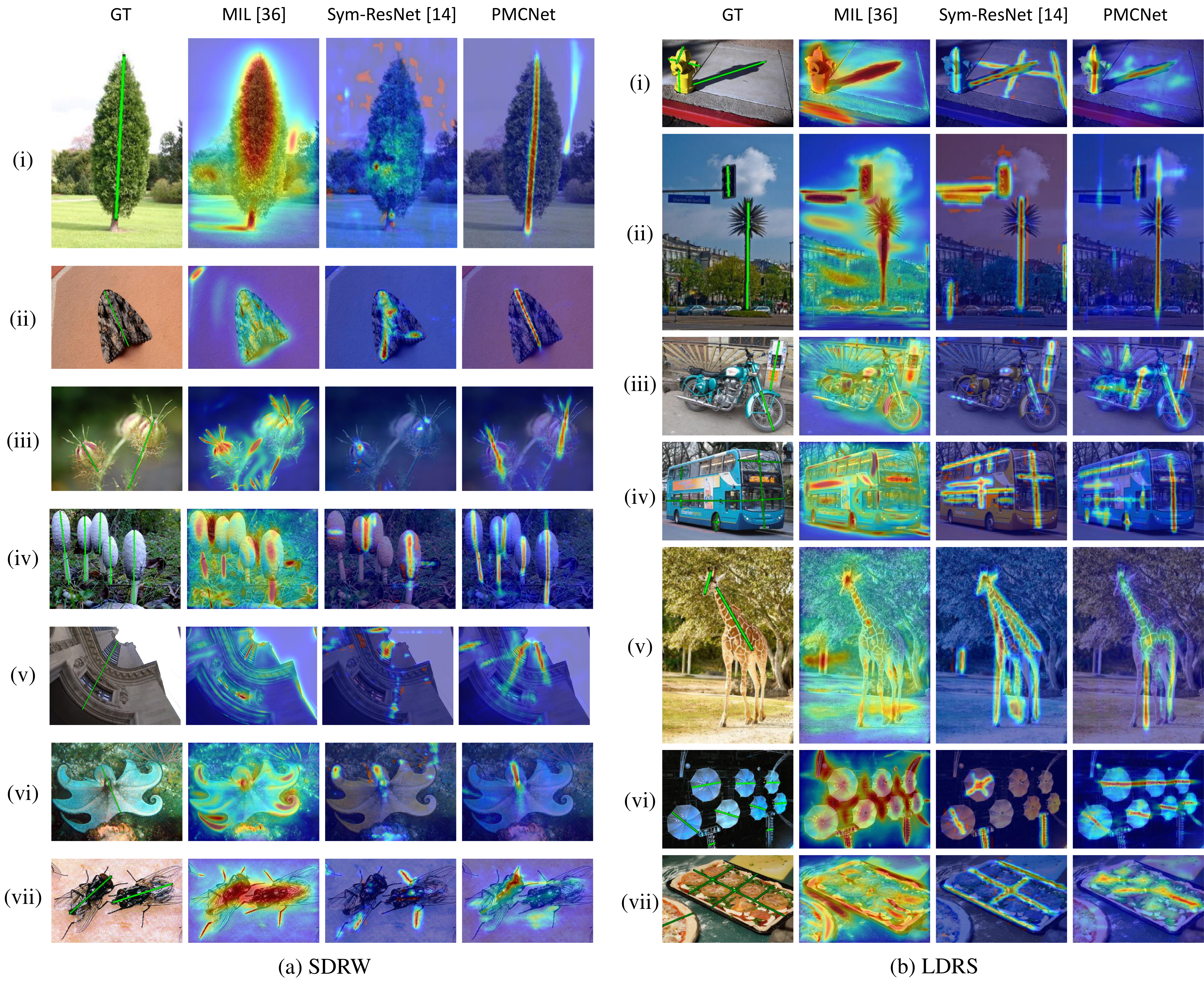}
    \caption{Qualitative results from the test splits of (a) SDRW and (b) LDRS datasets.}
    \label{fig:qual}
\end{figure*}

\subsection{Comparison with the state-of-the-art methods}
We compare our method with the state-of-the-art methods in~\Fig{pr_curve}. Our method achieves the F1 score of 68.8\% and outperforms the state-of-the-art method Sym-ResNet~\cite{funk2017beyond} with a large margin.
Notably, our model and LE~\cite{loy2006detecting} show robust performance against dense detection methods~\cite{funk2017beyond, shen2016object, teo2015detection, tsogkas2012learning}, which supports our claim that the reflection-equivariance of the feature descriptor is essential in detecting symmetry patterns.

The qualitative results are shown in~\Fig{qual}.
We compare our PMCNet with MIL~\cite{tsogkas2012learning} and Sym-ResNet~\cite{funk2017beyond} on (a) SDRW~\cite{liu2013symmetry} and (b) LDRS (Ours)  datasets.
SDRW exhibit well-defined symmetries while LDRS include deformed real-world symmetries.
MIL works well at the simple cases such as \Fig{qual}a-(i) and \Fig{qual}b-(i),
but it mostly fails with the complicated and textured objects.
Sym-ResNet fails by missing the apparent symmetries in \Fig{qual}a-(i), adding irrelevant regions in \Fig{qual}b-(i), falsely detecting the nearby relevant regions in \Fig{qual}a-(ii) and \Fig{qual}b-(ii). 
In contrast, PMCNet successfully detects the well-defined symmetries in \Fig{qual}a-(ii-iv) and the complex cases in~\Fig{qual}b-(ii-iv) where the other methods fail. 
Furthermore, PMCNet even detects symmetries of hierarchical patterns in \Fig{qual}a-(v) and thin object in \Fig{qual}b-(v), which were missed in the ground-truth annotations. 
PMCNet fails at detecting the ambiguous symmetries in \Fig{qual}a-(vi), multi-angled symmetries in \Fig{qual}b-(vi), and the symmetries at multiple scales in \Fig{qual}a-(vii) and \Fig{qual}b-(vii). 

PMCNet remains some limitations. 
The discrete and sparse bins of the polar descriptors might not fully represent the rotational freedom of the symmetry axis. 
Also, a single-level PMC with fixed polar window sizes is not capable of detecting the complex multi-scale symmetries.
Improving the model considering aforementioned issues and collecting the larger dataset will be interesting for future work.

\section{Conclusion}
We have introduced the polar matching convolution (PMC) to discover reflection symmetry patterns.
It effectively learns to detect symmetry patterns by leveraging the high-dimensional matching kernel with the relational descriptors, achieving the state of the art on the SDRW reflection symmetry benchmark.
The ablation studies demonstrate the effectiveness of the components of PMCNet - a polar feature pooling, a self-similarity encoding, and a systematic kernel design for axes of different angles. 
Further research on this direction can benefit different symmetry detection tasks and a wide range of problems related to symmetries. 

\smallbreak
\noindent \textbf{Acknowledgements.}
This work was supported by Samsung Advanced Institute of Technology (SAIT), the NRF grant (NRF-2017R1E1A1A01077999), and the IITP grant (No.2019-0-01906, AI Graduate School Program - POSTECH) funded by Ministry of Science and ICT, Korea.\looseness=-1

\nocite{*}
{\small
\bibliographystyle{ieee_fullname}
\bibliography{egbib}
}
\clearpage
\appendix

\setcounter{section}{0}
\setcounter{figure}{0}
\setcounter{table}{0}

\renewcommand{\thesection}{A.\arabic{section}}
\renewcommand{\thefigure}{a.\arabic{figure}}
\renewcommand{\thetable}{a.\arabic{table}}

\section*{Appendix}

\section{Synthetic augmentation}

\begin{figure}[h!]
    \centering
    \includegraphics[width=0.37\textwidth]{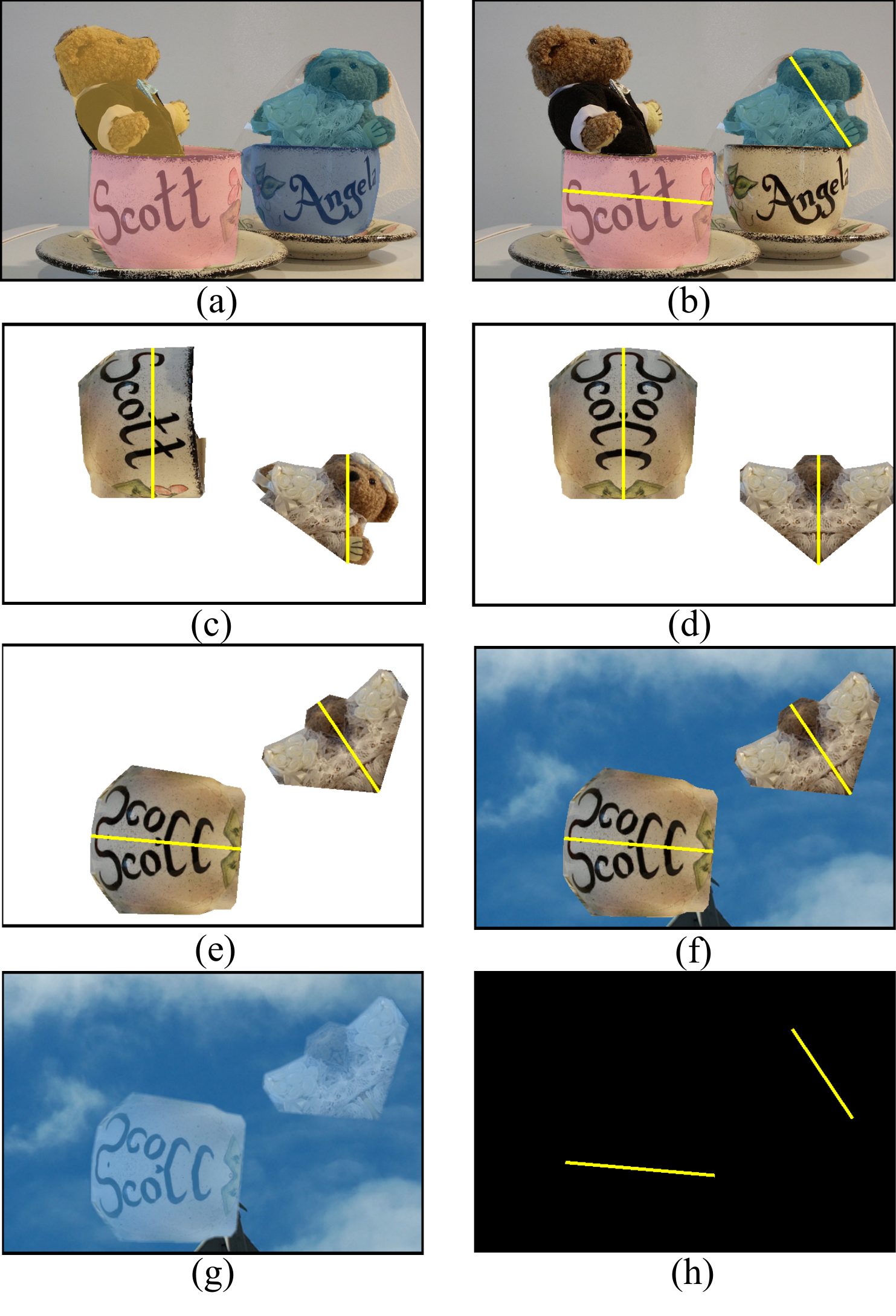}
    \caption{Synthesis process. (a) Given an image and its polygon annotations of instances from COCO, (b) we select some instances that show the maximum length of symmetries using depth-first-search (DFS) after synthesization. To construct symmetries in arbitrary angles (c) we rotate them in random degrees, (d) flip the left part of an vertical axis to the right and (e) rotate them back to the reverse direction with the same amount of angles. (f) Finally, we composite the foreground symmetries onto a randomly sampled background image from COCO. The alpha-blended version (g) and the axes heatmap (h) are used for training.}
    \label{fig:synthetic_generation}
\end{figure}

In this section, we provide a step-by-step process of synthesizing the symmetry images and their axes in Fig.~\ref{fig:synthetic_generation} and Alg.~\ref{alg1:synthdata}.

\makeatletter
\newcommand{\removelatexerror}{\let\@latex@error\@gobble}
\makeatother

\newcommand{\myalgorithm}{%
\begingroup
\removelatexerror%
\begin{algorithm}[ht]
\SetAlgoLined
\SetKwInOut{Input}{Input}\SetKwInOut{Output}{Output}
\Input{foreground and background images ($I_{\mathrm{fg}}$~and~$I_{\mathrm{bg}}$) and polygon instance annotations $\Omega_0$ of $I_{\mathrm{fg}}$}
\Output{heatmap of symmetry axis $Y$ and blended image $I_{\mathrm{sym}}$}
\BlankLine

$\Psi_0$: initial list of angles sampled within $[(0, \frac{1}{n}\pi),...,(\frac{n-1}{n}\pi, \pi)]$ given $n=2$\\
$\Psi$: selected list of angles from $\Psi_0$ \\
$L$: randomly sampled list of vertical axes \\
$\Omega$: selected list of annotations from $\Omega_0$\\

\BlankLine
$\Omega_0 \gets$ SortInstancesByArea($\Omega_0$)\\
$\Omega, \mathrm{L}, \Psi \gets$ EmptyLists\\
\For {$\omega_0 \in \Omega_0$}{
    \For {$\psi \in \Psi_0$}{
        $\omega \gets$ RotatePolygon($\omega_0$, $\psi$)\\
        /* sample vertical symmetry axis $l$ between 1/3 and 2/3 positions of the instance */ \\
        $\omega \gets$ FlipLeftPart($\omega$, vertical axis=$l$)\\
        $\omega \gets$ RotatePolygon($\omega$, $-\psi$)\\
        $\Omega$.append($\omega$), $\mathrm{L}$.append($\l$), $\Psi$.append($\psi$)\\
    }
}

\BlankLine
/* Find the best combination that maximize the length of the symmetry axes by DFS*/\\
$\Omega, \mathrm{L}, \Psi \gets$ BestSymmetryByDFS($\Omega$, $\mathrm{L}$, $\Psi$)\\
\BlankLine
/* Composite symmetry images with the same process that polygons are transformed. */\\
$I_{\mathrm{symfg}} \gets$ CompositeSymmetryFG($I_{\mathrm{fg}}$, $\Omega$, $\mathrm{L}$, $\Psi$)\\
$I_{\mathrm{sym}} \gets$ AlphaBlendwithBG($I_{\mathrm{symfg}}$, $I_{\mathrm{bg}}$, $\Omega'$)\\
\BlankLine
/* Composite symmetry axes by drawing and rotating vertical symmetric axes. */\\
$Y \gets$ DrawRotateVerticalAxes($\mathrm{L}$, $\Psi$)\\
\Return{$Y$, $I_{\mathrm{sym}}$}

\caption{Synthesize Data from COCO~\cite{lin2014microsoft}}
\label{alg1:synthdata}
\end{algorithm}
\endgroup}

\myalgorithm

\begin{figure*}[ht]
    \centering
    \includegraphics[width=0.75\textwidth]{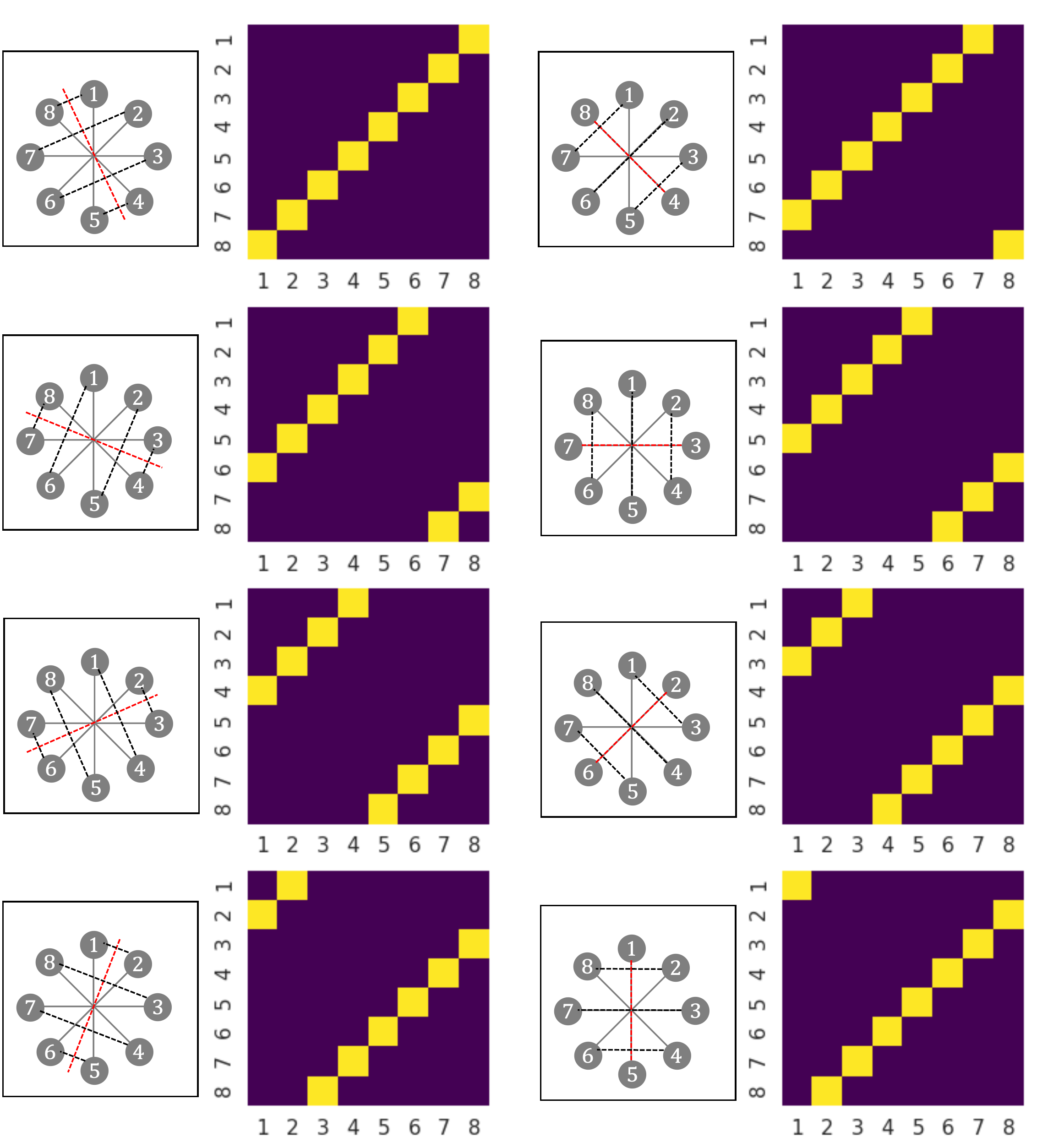}
    \caption{A visualization of the valid indices of the reflective matching kernel. Target axis (left) and corresponding kernel (right) are placed side to side. The kernel is binary where the bright yellow indicates \textit{valid}.}
    \label{fig:kernel_ours}
\end{figure*}

\section{Reflective matching kernel}
We show that the \textit{axis-aware} characteristic improves the kernel in Tab. 2 of the main paper. Although we illustrated the examples of the valid indices in the Fig.3 and Fig.4 of the main paper, we show the visualization of the valid indices for every candidate axes in ~\Fig{kernel_ours}.
Note that $N_\mathrm{axi}$ candidate axes for the $N_\mathrm{axi}$ points have two cases: (1) axes crossing the points and (2) axes lying between the points.
\begin{table}
\centering
\setlength\tabcolsep{4pt}
\centering
\resizebox{0.5\textwidth}{!}{%
\begin{tabular}{|ccc|ccc|} \hline
 \multicolumn{3}{|c|}{train dataset} & \multicolumn{3}{c|}{test dataset}  \\  \hline
 LDRS-\textit{synth} & SDRW + LDRS & NYU 
 & SDRW & LDRS & NYU \\ \hline \hline
 \checkmark & \checkmark &  & 63.4 & 34.8 & \textbf{62.1} \\
\hline
\checkmark & \checkmark & \checkmark & \textbf{68.8} & \textbf{37.3} & - \\ \hline
\end{tabular} 
}
\caption{Experimental results on the training sets with/without NYU~\cite{ConvSymm2016}.}
\label{tab:nyu}
\end{table}

\section{Experimental results on NYU dataset}
Since NYU~\cite{ConvSymm2016} does not distinguish the train and the test splits, we use them all for our training set in the main paper.
To investigate the effectiveness of the NYU dataset, we conduct the experiments by separating them from our training sets (\Tbl{nyu}).
Training with the NYU dataset offers improvements on F-score for the SDRW and the LDRS test datasets.
When testing the NYU dataset, our method achieves the F1-score of 62.1\%.

\end{document}